\newif\ifshowkorean
\newcommand{\KO}[1]{\ifshowkorean #1\fi}
\newcommand{\EN}[1]{\ifshowkorean\else #1\fi}

\documentclass[letterpaper, 10 pt, conference]{ieeeconf}  

\IEEEoverridecommandlockouts                              

\usepackage{amsmath} 
\usepackage{amssymb}  

\usepackage{iftex}  
\usepackage{graphicx} 
\usepackage{soul}  
\usepackage{silence}
\usepackage[caption=false]{subfig}
\usepackage{multirow}
\usepackage{siunitx}  
\usepackage{mathtools}  
\usepackage{mathrsfs}  
\usepackage{nameref}  
\usepackage[hidelinks]{hyperref}
\DeclarePairedDelimiterX{\kldivx}[2]{[}{]}{%
  #1\;\delimsize\|\;#2%
}
\newcommand{\kldiv}{D_{\mathrm{KL}}\kldivx}

\newcommand{\modalfont}[1]{\mathsf{#1}}
\ifxetex
    \usepackage{fontspec}

    \makeatletter
    \KO{
        \usepackage[AutoFallBack=true,CJKspace=true]{xeCJK}
        
        \setCJKmainfont[
        AutoFakeSlant,
        UprightFont=*-Regular.otf,
        BoldFont=*-Bold.otf,
        ]{MaruBuri}
    
        \setCJKsansfont[
        AutoFakeSlant,
        UprightFont=*-Regular.otf,
        BoldFont=*-Bold.otf,
        ]{MaruBuri}
    
        \setCJKmonofont[
        AutoFakeSlant,
        UprightFont=*-Regular.otf,
        BoldFont=*-Bold.otf
        ]{NotoSansMonoCJKkr}
    
        \setCJKfallbackfamilyfont{\CJKrmdefault}[AutoFakeSlant]{NotoSerifCJKjp-Light.otf}
        \setCJKfallbackfamilyfont{\CJKsfdefault}[AutoFakeSlant]{NotoSansCJKjp-Regular.otf}

        \def\normalsize{\@setfontsize{\normalsize}{9}{12pt}}%
    }
    \makeatother
\fi

\KO{
    \title{\LARGE \bf {거울-표식 태스크에서의 self-prior를 이용한 능동적 추론}}
}
\EN{
    \title{\LARGE \bf {Active Inference with a Self-Prior in the Mirror-Mark Task
    }}
}

\author{Dongmin Kim$^{1}$, Hoshinori Kanazawa$^{1,2}$, Yasuo Kuniyoshi$^{1,2}$
\thanks{This paper was supported by JST PRESTO (Grant Number JPMJPR23S4), JSPS KAKENHI (Grant Number JP25K24741), and the Next Generation Artificial Intelligence Research Center (AI Center), The University of Tokyo.}
\thanks{$^{1}$Laboratory for Intelligent Systems and Informatics, Graduate School of Information Science and Technology, The University of Tokyo, 7-3-1 Hongo, Bunkyo-ku, Tokyo, Japan (e-mail: {\tt\small\{d-kim, kanazawa, kuniyosh\}@isi.imi.i.u-tokyo.ac.jp})}
\thanks{$^{2}$Co-corresponding authors}
\thanks{†To support reproducibility, we release a simplified robot-based implementation of our method, while the full infant simulation environment remains private.}
}

\begin{document}

\maketitle
\thispagestyle{empty}
\pagestyle{empty}

\begin{abstract}
\KO{
    거울 자기 인식 테스트는 거울에서만 보이는 표식을 자신의 몸에서 만지는 행동을 평가하는 실험으로, 자기 인식 능력의 지표로 널리 사용된다. 본 연구에서는 단일 메커니즘인 self-prior만으로, 외부 보상 없이 이 행동이 자발적으로 발생하는 계산론적 모델을 보인다.
    트랜스포머로 구현된 Self-prior는 익숙한 다감각 경험의 밀도를 학습하며, 새로운 표식이 나타나면 이 분포와의 불일치가 능동적 추론을 통해 표식을 향한 행동을 유발한다.
    시뮬레이션된 영아는 촉각 없이 시각과 고유수용감각만을 사용하여 거울 속 자신의 얼굴에 붙은 스티커를 발견하고, 명시적 지시 없이 약 70\%의 확률로 스티커를 제거하였다.
    스티커 제거 이후 기대 자유 에너지가 감소하였는데, 이는 self-prior가 익숙한 자기 관련 상태를 선호하는 내부 기준으로 작동한다는 해석과 부합한다.
    또한 교차 모달 재구성을 통해 self-prior가 시각-고유수용감각 연관을 포착하여 확률적 body schema로 기능함을 보였다.
    본 결과는 거울 검사에서 관찰되는 핵심 행동에 대한 간결한 계산론적 설명을 제공하며, 자유 에너지 원리가 자기 인식의 발달적 기원을 탐구하는 가설이 될 수 있음을 시사한다. 코드$^\text{†}$는 다음에서 제공된다: \href{https://github.com/kim135797531/self-prior-mirror}{https://github.com/kim135797531/self-prior-mirror}.
}\EN{
The mirror self-recognition test evaluates whether a subject touches a mark on its own body that is visible only in a mirror, and is widely used as an indicator of self-awareness.
In this study, we present a computational model in which this behavior emerges spontaneously through a single mechanism, the self-prior, without any external reward.
The self-prior, implemented with a Transformer, learns the density of familiar multisensory experiences; when a novel mark appears, the discrepancy from this learned distribution drives mark-directed behavior through active inference.
A simulated infant, relying solely on vision and proprioception without tactile input, discovered a sticker placed on its own face in the mirror and removed it in approximately 70\% of cases without any explicit instruction.
Expected free energy decreased after sticker removal, consistent with the self-prior acting as an internal criterion that favors familiar self-related states.
Cross-modal reconstruction further demonstrated that the self-prior captures visual--proprioceptive associations, functioning as a probabilistic body schema.
These results provide a concise computational account of the key behavior observed in the mirror test and suggest that the free energy principle can serve as a unifying hypothesis for investigating the developmental origins of self-awareness. Code$^\text{†}$ is available at: \href{https://github.com/kim135797531/self-prior-mirror}{https://github.com/kim135797531/self-prior-mirror}.
}
\end{abstract}

\section{INTRODUCTION}

\KO{
    거울 자기 인식 테스트는 동물이 거울 속 이미지를 자기 자신으로 인식하는지를 평가하는 실험으로, 자기 인식 능력의 핵심 지표로 널리 사용된다.
    Gallup은 1970년에 침팬지가 거울 속 자신의 얼굴에 표시된 마크를 만지는 행동을 통해 자기 인식을 입증하였다~\cite{gallup_chimpanzees_1970}.
    이후 Amsterdam은 인간 영아가 약 18--24개월경에 거울 자기 인식 능력을 획득한다는 것을 rouge 테스트를 통해 밝혔다~\cite{amsterdam_mirror_1972}.
}
\EN{
The mirror self-recognition test evaluates whether an animal recognizes its reflection as itself, serving as a key indicator of self-awareness.
Gallup demonstrated in 1970 that chimpanzees exhibit self-recognition by touching marks placed on their faces while viewing their mirror reflection~\cite{gallup_chimpanzees_1970}.
Subsequently, Amsterdam showed through the rouge test that human infants acquire mirror self-recognition ability around 18--24 months of age~\cite{amsterdam_mirror_1972}.
}

\KO{
    그러나 계산론적 관점에서, 거울 속 이상을 감지한 뒤 왜 자신의 몸을 향한 행동이 자발적으로 발생하는지는 여전히 충분히 설명되지 않았다.
    기존의 거울 자기 인식 계산 모델들은 주로 시각적 이상의 위치를 운동 명령으로 명시적으로 변환하거나, 자타 구별에 필요한 중간 표현을 별도로 학습하는 방식에 의존했다~\cite{hoffmann_robot_mirror_2021, lanillos_robot_self_other_2020}.
    이러한 접근은 거울 속 표식과 행동의 연결을 구현하는 데 유용하지만, 표식 제거 행동을 유발하는 내부 기준 자체는 대개 외부에서 설계된 모듈이나 특징 표현에 의존한다.
    핵심 질문은, 외부 보상이나 명시적 목표 없이 에이전트 자신의 경험만으로 거울 속 표식을 향한 행동이 어떻게 창발할 수 있는가이다.
    }
    \EN{
        
From a computational perspective, however, it remains insufficiently explained why detecting an anomaly in the mirror should spontaneously lead to action directed toward one's own body.
Existing computational models of mirror self-recognition have typically relied on explicitly transforming the location of a visual anomaly into motor commands or on learning dedicated intermediate representations for self-other distinction~\cite{hoffmann_robot_mirror_2021, lanillos_robot_self_other_2020}.
These approaches are useful for implementing the link between a mark in the mirror and action, but the internal criterion that drives mark-removal behavior is still often supplied by externally designed modules or feature representations.
The central question is how mark-directed behavior can emerge from the agent's own experience alone, without external reward or an explicit goal.

}

\KO{
행동 선택에 자유 에너지 원리~\cite{friston_free-energy_2010}를 적용하는 능동적 추론~\cite{friston_active_2016}은 이 질문에 대한 원리적 해답을 제안한다.
능동적 추론에서 에이전트는 기대 자유 에너지를 최소화하는 방향으로 행동을 선택하므로, 외부 보상이 없더라도 감각적 이상 자체가 행동을 촉발하는 내적 동인이 될 수 있다.
본 연구에서는 이 프레임워크 위에 self-prior를 도입한다: self-prior는 에이전트가 평소 경험해 온 ``자기다운'' 다감각 패턴의 분포를 학습하고, 현재 감각이 그 익숙한 분포에서 얼마나 벗어나는지를 평가하는 내부 모델이다~\cite{kim2025emergence}.
이러한 self-prior는 에이전트 자신의 익숙한 다감각 경험의 밀도를 모델링한다는 점에서, 탐색 효율에 초점을 맞춘 기존의 내발적 동기 연구와 구별된다.
나아가 self-prior는 현재 감각과 익숙한 자기 상태 사이의 불일치를 직접 평가함으로써, 거울 속 표식의 위치를 명시적으로 계산하지 않고도 행동을 유도하는 계산적 기준이 될 수 있다.
}
\EN{

Active inference~\cite{friston_active_2016}, which implements the free energy principle~\cite{friston_free-energy_2010} for action selection, suggests a principled answer to this question.
Because agents under active inference select actions that minimize expected free energy, a sensory anomaly can itself become an intrinsic drive for action, even without external reward.
Building on this framework, we introduce the self-prior: an internal model that learns the distribution of multisensory patterns characteristic of the agent and evaluates how far the current sensation deviates from that familiar distribution~\cite{kim2025emergence}.
The self-prior differs from prior intrinsic-motivation approaches that mainly focus on improving exploration efficiency, because it models the density of the agent's own familiar multisensory experiences.
Moreover, by directly evaluating the mismatch between the current sensation and a familiar self state, the self-prior can serve as a computational criterion that guides behavior without explicitly computing the mark location in the mirror.
 }

\begin{figure}[t]
    \centering
    \subfloat[]{\includegraphics[width=0.49\columnwidth]{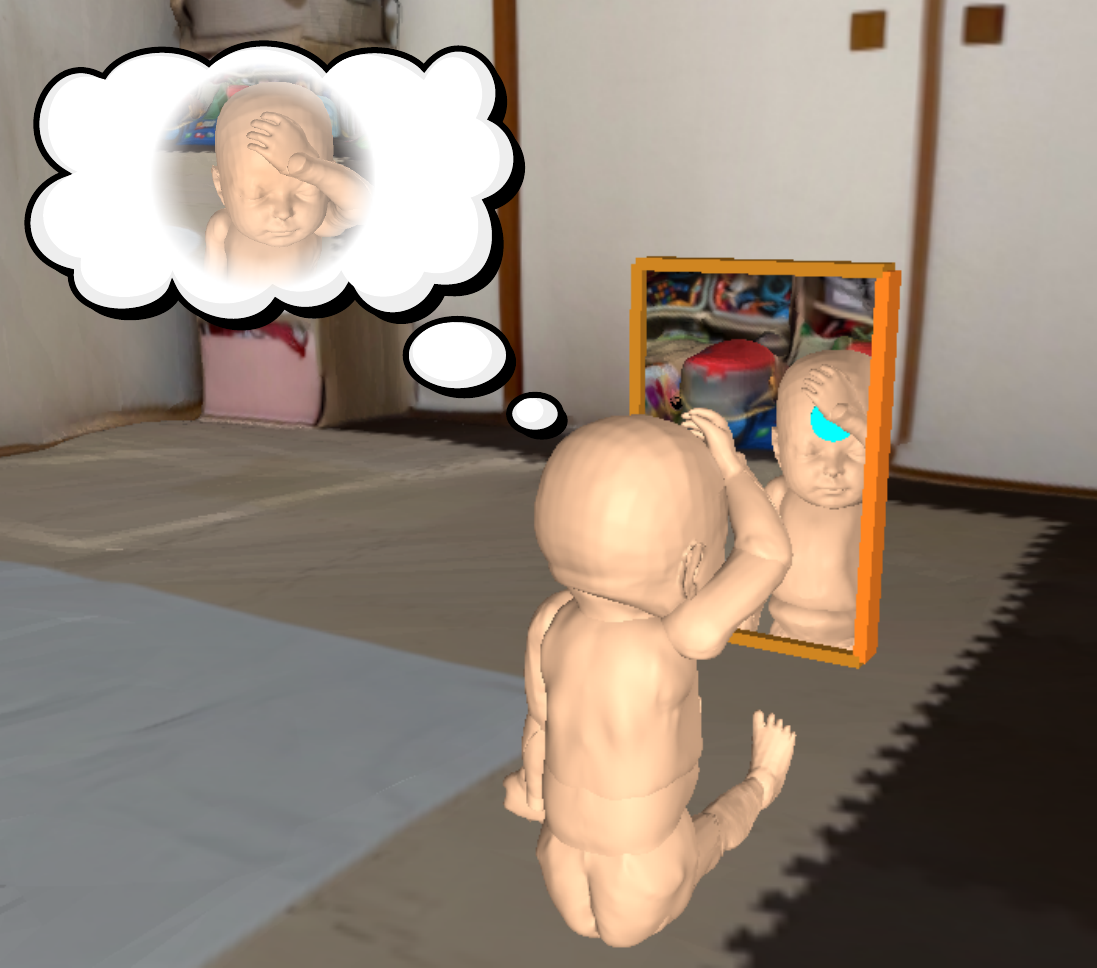}\label{fig:fig1a}}
    \hfil
    \subfloat[]{\includegraphics[width=0.43\columnwidth]{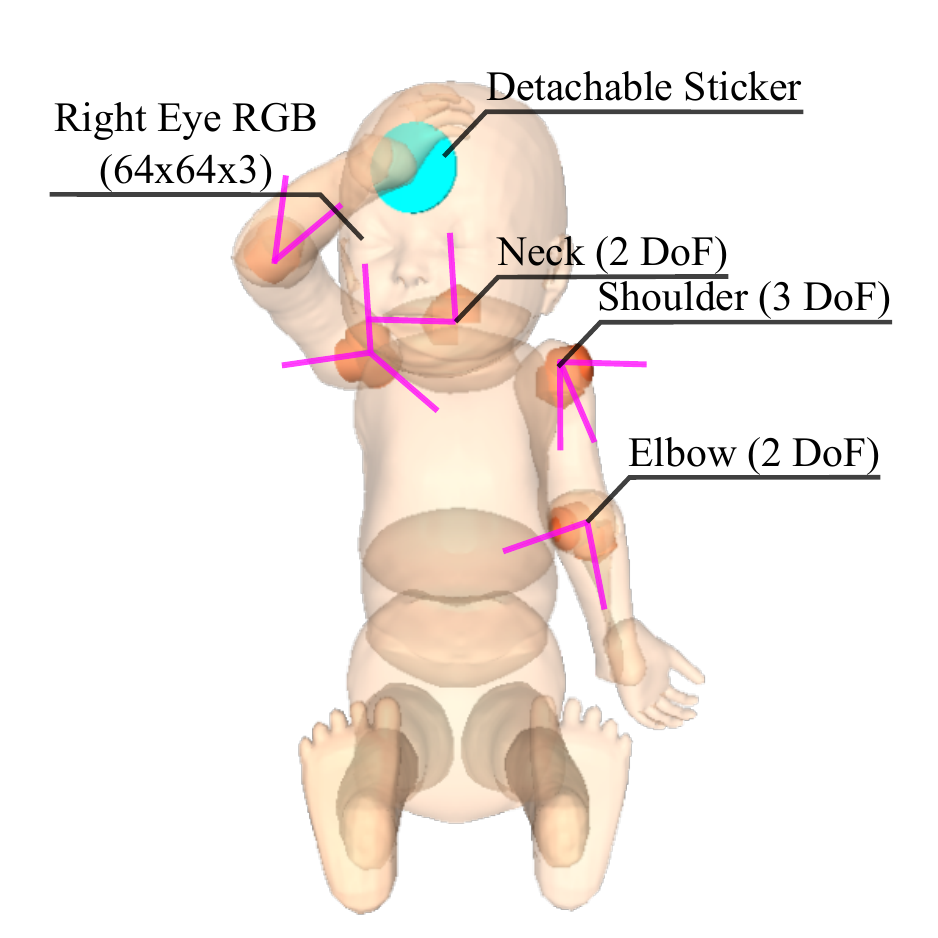}\label{fig:fig1b}}
    \par\vspace{1em}
    \subfloat[]{\includegraphics[width=\columnwidth]{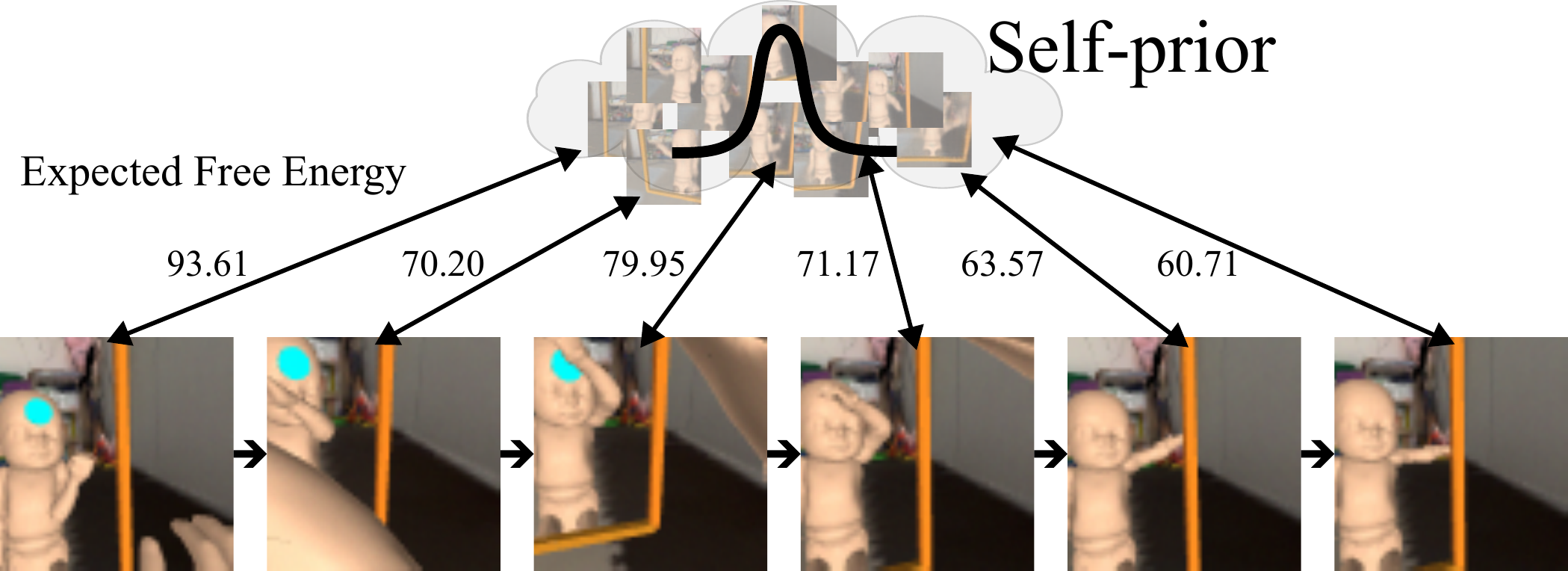}\label{fig:fig1c}}
    \caption{
    \KO{실험 개요. (a) 거울 앞에 앉은 시뮬레이션된 영아는 거울 속 자신의 모습을 인식하고 스티커를 제거한다. (b) 영아는 시각과 관절 각도의 고유수용감각을 받는다. 각 관절은 가상의 모터에 의해 제어된다. (c) 익숙한 감각 경험의 밀도를 모델링한 self-prior와의 비교를 통해 기대 자유 에너지를 계산한다. 정책 네트워크는 기대 자유 에너지를 최소화하는 행동을 선택하도록 학습된다.}
    \EN{Experimental overview. (a) A simulated infant sitting in front of a mirror recognizes its reflection and removes the sticker. (b) The infant receives visual input and proprioceptive input of joint angles. Each joint is controlled by a virtual motor. (c) Expected free energy is computed by comparing with the self-prior that models the density of familiar sensory experiences. The policy network is trained to select actions that minimize expected free energy.}
    }
    \label{fig:overall}
\end{figure}

\KO{
    본 연구에서는 트랜스포머 기반 self-prior와 능동적 추론을 결합하여, 시뮬레이션된 영아가 거울 속 자신의 얼굴에 붙은 스티커를 발견하고 제거하는 과정을 모델링한다~(Fig.~\ref{fig:overall}).
    Self-prior는 스티커가 없는 일상적인 상황에서 학습되며, 스티커가 부착되면 에이전트는 익숙한 자신의 모습과 현재 관찰 사이의 불일치를 감지한다.
    이에 따라 에이전트는 원시 픽셀과 고유수용감각을 end-to-end로 처리하면서, 기대 자유 에너지를 최소화하는 방향으로 스티커를 향해 손을 뻗는다.
    중요한 점은, 에이전트가 촉각 없이 오직 시각과 고유수용감각만을 사용한다는 것으로, 이는 거울 속 이미지를 자신의 몸과 연결하는 내부 기준이 실제 행동 생성에 사용됨을 보여준다.
    본 연구는 외부 보상 없이도 self-prior가 거울 검사에서 나타나는 표식 지향 행동의 내부 기준으로 작동할 수 있음을 보이며, 자기 인식의 발달적 기원을 탐구하기 위한 계산론적 토대를 제공한다.
}
\EN{
In this study, we model the process by which a simulated infant discovers and removes a sticker on its own face in a mirror by combining a Transformer-based self-prior with active inference~(Fig.~\ref{fig:overall}).
The self-prior is trained in everyday situations without stickers, and when a sticker is attached, the agent detects a mismatch between its familiar appearance and the current observation.
The agent then processes raw pixels and proprioception end-to-end and reaches toward the sticker in a way that minimizes expected free energy.
Crucially, the agent relies solely on vision and proprioception without tactile feedback, showing that an internal criterion linking the mirrored image to its own body can actually be used to generate behavior.
This study demonstrates that, even without external rewards, the self-prior can serve as an internal criterion for the mark-directed behavior observed in the mirror test, offering a computational basis for investigating the developmental origins of self-awareness.
}

\section{METHOD}

\subsection{Simulation Environment}

\KO{
    우리는 MuJoCo 물리 시뮬레이션~\cite{todorov_mujoco_2012} 기반의 영아 시뮬레이션 플랫폼 EMFANT~\cite{kim_simulating_2022} 위에, 거울 앞에 앉은 영아 에이전트를 구성하였다.
    영아 모델은 목 \qty{2}{DoF}, 각 팔의 어깨 \qty{3}{DoF}, 팔꿈치 \qty{2}{DoF}로 구성되어 총 \qty{12}{DoF}를 갖고 있다.
    각 관절에는 양방향으로 최대 \qty{1}{\N\m}의 토크로 제어되는 가상의 모터가 부착되어 있다.
    거울의 크기는 가로 \qty{20}{\cm} 세로 \qty{30}{\cm}이며, 오른쪽 눈과 거울 평면 사이의 초기 거리는 약 \qty{17.7}{\cm}이다.

    에이전트는 두 가지 감각 양상을 통해 환경을 관찰한다.
    시각 입력 $o_t^\modalfont{V}$는 오른쪽 눈의 RGB 이미지이다.
    고유수용감각 입력 $o_t^\modalfont{P} \in \mathbb{R}^{12}$는 각 관절 각도를 실수 값으로 제공하며, 관절 가동 범위 내로 클리핑된다.
    거울이 없어도 촉각만으로 표식을 제거하는 것을 방지하기 위해 촉각 입력은 제공하지 않았다.
    운동 출력 $a_t \in \mathbb{R}^{12}$는 각 관절의 연속 제어 신호이다.
}
\EN{
    We constructed an infant agent seated in front of a mirror on EMFANT\footnote{\href{https://www.isi.imi.i.u-tokyo.ac.jp/public/emfant}{https://www.isi.imi.i.u-tokyo.ac.jp/public/emfant}}~\cite{kim_simulating_2022}, an infant simulation platform based on the MuJoCo physics simulator~\cite{todorov_mujoco_2012}.
    The infant model has \qty{12}{DoF} in total, with \qty{2}{DoF} in the neck, \qty{3}{DoF} in each shoulder, and \qty{2}{DoF} in each elbow.
    Each joint is actuated by a virtual motor controlled bidirectionally with a maximum torque of \qty{1}{\N\m}.
    The mirror measures \qty{20}{\cm} by \qty{30}{\cm}, and the initial distance between the right eye and the mirror plane is approximately \qty{17.7}{\cm}.

    The agent observes the environment through two sensory modalities.
    Visual input $o_t^\modalfont{V}$ is provided as an RGB image from the right eye.
    Proprioceptive input $o_t^\modalfont{P} \in \mathbb{R}^{12}$ provides each joint angle as a real-valued signal and is clipped to the allowable range of motion.
    Tactile input was not provided to prevent sticker removal based on touch alone, even without the mirror.
    Motor output $a_t \in \mathbb{R}^{12}$ consists of continuous control signals for each joint.
}

\subsection{Model Architecture}

\begin{figure}[t]
    \centering
    \includegraphics[width=0.8\columnwidth]{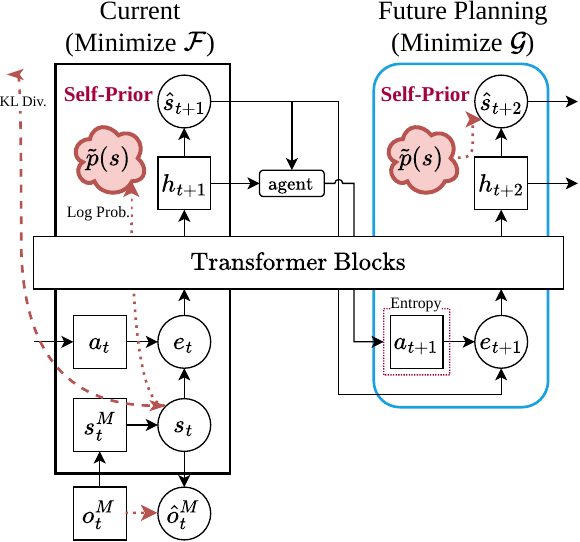}
    \caption{
    \KO{Self-prior를 이용한 능동적 추론 프레임워크. Self-prior는 관찰이 익숙한 패턴과 얼마나 일치하는지를 평가한다. World model은 현재 관찰로부터 잠재 상태를 추론하고, 정책에 따른 미래 궤적을 상상한다. 기대 자유 에너지는 상상된 잠재 상태와 self-prior와의 불일치로부터 계산된다.}
    \EN{Active inference framework with self-prior. The self-prior evaluates how well the current observation matches familiar patterns. The world model infers latent states from observations and imagines future trajectories under the policy. Expected free energy is computed from the mismatch between imagined latent states and the self-prior.}
    }
    \label{fig:model_architecture}
\end{figure}

\KO{
    본 연구는 거울 자기 인식을 능동적 추론의 관점에서 접근한다.
    이 프레임워크를 구현하기 위해 세 가지 모듈이 필요하다: 변분 자유 에너지 최소화로 구축되는 world model, 익숙한 감각 경험의 밀도를 모델링하는 self-prior, 그리고 기대 자유 에너지 최소화로 구축되는 정책 네트워크이다 (Fig.~\ref{fig:model_architecture}).
    구현은 DreamerV3~\cite{hafner_mastering_2023}와 유사한 학습 절차를 사용하면서 트랜스포머 기반 시계열 모델을 사용하는 STORM~\cite{zhang_storm_2023}을 기반으로 하며, 심층 신경망을 활용하여 능동적 추론의 핵심 계산을 고차원 문제로 확장한다~\cite{millidge2020deep}.
    본 연구에서는 외부 보상 대신 self-prior가 행동의 내부 기준을 제공한다.
}
\EN{
    This study approaches mirror self-recognition from the perspective of active inference.
    Three modules are required to implement this framework: a world model built on variational free energy minimization, a self-prior that models the density of familiar sensory experiences, and a policy network built on expected free energy minimization (Fig.~\ref{fig:model_architecture}).
    The implementation builds on STORM~\cite{zhang_storm_2023}, which uses a DreamerV3-like training pipeline~\cite{hafner_mastering_2023} with Transformer-based sequence modeling, and extends the core computations of active inference to high-dimensional problems via deep neural network approximations~\cite{millidge2020deep}.
    In our setting, the internal criterion for behavior is provided by the self-prior rather than by external reward.
}

\subsubsection{World Model}

\KO{
    World model은 관찰 정보의 압축을 위한 인코더/디코더와, 시계열 예측을 위한 트랜스포머로 구성된다.
    관찰 인코더/디코더는 시각 입력 $o^\modalfont{V}_t$와 고유수용감각 입력 $o^\modalfont{P}_t$를 받아 one-hot 통합 잠재 상태 $s_t$의 분포를 생성하는 Categorical VAE이다:
}
\EN{
    The world model consists of an encoder/decoder for observation compression and a Transformer for temporal prediction.
    The observation encoder/decoder is a categorical VAE that takes visual input $o^\modalfont{V}_t$ and proprioceptive input $o^\modalfont{P}_t$ and produces the distribution over the unified one-hot latent state $s_t$:
}

\begin{equation*}
    \begin{alignedat}{2}
        &\forall \modalfont{M} &&\in \{\modalfont{V},\modalfont{P}\} \\
        \text{Observation encoder:}~ & s^\modalfont{M}_t &&= g^\modalfont{M}_{\phi}(o^\modalfont{M}_t) \\
        \text{Observation mixer:}~ & \bar{s}_t &&= g^s_{\phi}(s^\modalfont{V}_t, s^\modalfont{P}_t) \\
        \text{Observation posterior:}~ & s_t &&\sim q_{\phi}(s_t \mid \bar{s}_t) \\
        \text{Inverse mixer:}~ & \hat{\bar{s}}^\modalfont{M}_t &&= g^{'\modalfont{M}}_{\phi}(s_t) \\
        \text{Observation decoder:}~ & \hat{o}^\modalfont{M}_t &&\sim p^\modalfont{M}_{\phi}(\hat{o}
        ^\modalfont{M}_t \mid \hat{\bar{s}}^\modalfont{M}_t)
    \end{alignedat}
\end{equation*}

\KO{
    Posterior layer는 $\bar{s}_t$를 각 $s_t^{(k)}$가 32개 클래스의 one-hot 벡터인 이산 잠재 상태 $s_t = (s_t^{(1)}, \ldots, s_t^{(32)})$로 인코딩한다.
    $s_t$를 결정론적 역믹서 $g^{'\modalfont{M}}_\phi$를 통해 각 modal의 임베딩의 추정값 $\hat{\bar{s}}^\modalfont{M}_t$로 되돌린 뒤, 이로부터 디코더는 관찰을 복원한다.

    액션 믹서는 잠재 상태 $s_t$와 행동 $a_t$를 시퀀스 모델의 입력 $e_t$로 변환한다.
    Dynamics predictor는 단일 linear layer로 구성되어 $h_{t+1}$로부터 categorical 분포인 다음 잠재 상태 $\hat{s}_{t+1}$를 생성한다:
}
\EN{
    The posterior layer encodes $\bar{s}_t$ into the discrete latent state $s_t = (s_t^{(1)}, \ldots, s_t^{(32)})$, where each $s_t^{(k)}$ is a one-hot vector over 32 classes.
    The latent state $s_t$ is mapped back through the deterministic inverse mixer $g^{'\modalfont{M}}_\phi$ into the estimated embedding $\hat{\bar{s}}^\modalfont{M}_t$ of each modality, from which the decoder reconstructs the observation.

    For temporal prediction, the action mixer converts the latent state $s_t$ and action $a_t$ into the sequence-model input $e_t$.
    The dynamics predictor, a single linear layer, generates a categorical distribution over the next latent state $\hat{s}_{t+1}$ from $h_{t+1}$:
}

\begin{equation*}
    \begin{alignedat}{2}
        \text{Action mixer:}& ~\quad e_t &&= g^a_{\phi}(s_t, a_t) \\
        \text{Sequence model:}& ~ h_{2:T+1} &&= f_{\phi}(e_{1:T}) \\
        \text{Dynamics predictor:}& ~\quad \hat{s}_{t+1} &&\sim p_{\phi}(\hat{s}_{t+1} \mid h_{t+1})
    \end{alignedat}
\end{equation*}

\KO{
    원 STORM은 reward predictor 및 continuation predictor를 포함하지만, 본 연구에서는 self-prior가 자유 에너지 계산을 담당하고 에피소드 도중에 일찍 종료되는 경우도 없으므로 이 두 예측기는 제외하였다.
}
\EN{
    The original STORM includes a reward predictor and a continuation predictor; we omit both because the self-prior handles free energy computation and there is no early episode termination.
}

\subsubsection{Self-Prior}

\KO{
    Self-prior $\tilde{p}_\xi(s)$는 에이전트의 일상적인 감각 경험의 밀도를 잠재 공간에서 모델링하는 분포이다.
    능동적 추론에서 기대 자유 에너지의 선호 분포는 에이전트가 달성하려는 목표 상태를 정의하는데, Kim 등~\cite{kim2025emergence}은 이 선호 분포를 에이전트 자신의 경험으로부터 학습된 self-prior로 대체함으로써 목적론적 행동이 내발적으로 창발할 수 있음을 보였다.
    본 연구에서는 관찰 공간 $o$의 밀도 대신 잠재 공간 $s$의 밀도 $\tilde{p}_\xi(s)$를 사용한다.
    World model의 학습을 통해 잠재 상태 $s_t$는 관찰 $o_t$의 충분 통계량이 되도록 학습되므로, 잠재 공간에서의 밀도 평가가 관찰 공간에서의 밀도 평가를 근사할 수 있다.

    Self-prior는 GPT 유사 Transformer로 구현되어, 통합 잠재 상태의 결합 분포를 autoregressive하게 모델링한다.
    결합 분포 계산을 위해 BOS (Beginning-of-Sequence) 토큰을 추가하며, 각 이산 변수 $s^{(k)}$는 $32+1$개의 vocabulary를 가진 토큰으로 임베딩된다:
}
\EN{
    The self-prior $\tilde{p}_\xi(s)$ is a distribution that models the density of the agent's everyday sensory experiences in latent space.
    In active inference, the preferred distribution in expected free energy defines the goal states the agent seeks to achieve; Kim et al.~\cite{kim2025emergence} showed that replacing this distribution with a self-prior learned from the agent's own experience allows goal-directed behavior to emerge intrinsically.
    In this work we use the density $\tilde{p}_\xi(s)$ in latent space instead of the density in observation space.
    Because the latent state $s_t$ is trained to be a sufficient statistic of observation $o_t$ through world model learning, density evaluation in latent space approximates that in observation space.
    The self-prior is implemented as a GPT-like Transformer that autoregressively models the joint distribution of the unified latent state.
    A BOS (Beginning-of-Sequence) token is added for joint distribution computation, and each discrete variable $s^{(k)}$ is embedded as a token with a vocabulary of $32+1$:
}

\begin{equation*}
    \tilde{p}_\xi(s_t) = \prod_{k=0}^{32} \tilde{p}_\xi(s_t^{(k)} \mid s_t^{(<k)}),\quad s_t^{(0)} = \text{BOS}
\end{equation*}

\subsubsection{Policy and Value Networks}

\KO{
    정책 네트워크와 가치 네트워크는 world model의 잠재 상태 $s_t$와 hidden state $h_t$를 입력받아 행동 분포와 가치를 예측한다.
    정책 네트워크는 기대 자유 에너지를 최소화하는 행동 $a_t$을 샘플링하고, 가치 네트워크는 GAE $\lambda$-return 계산을 위한 baseline을 제공한다:
}
\EN{
    The policy and value networks take the world model's latent state $s_t$ and hidden state $h_t$ as input and predict the action distribution and value.
    The policy network samples actions $a_t$ that minimize expected free energy, and the value network provides a baseline for GAE $\lambda$-return computation:
}

\begin{equation*}
    \begin{alignedat}{2}
        \text{Policy:}& ~\quad a_t &&\sim \pi_\theta(a_t \mid s_{t}, h_{t}) \\
        \text{Expected utility (Value):}& ~\quad v_t &&= V_\psi(s_t, h_t)
    \end{alignedat}
\end{equation*}

\KO{
    정책 네트워크는 tanh 변환된 정규분포를 출력하며, 가치 네트워크는 DreamerV3의 symlog two-hot 인코딩을 사용한다.
}
\EN{
    The policy network outputs a tanh-transformed normal distribution, and the value network uses the symlog two-hot encoding of DreamerV3.
}

\subsection{Active Inference with Self-Prior}

\KO{
    자유 에너지 원리에서 에이전트는 관찰을 설명하기 위해 변분 자유 에너지를 최소화하고, 행동 선택을 위해 기대 자유 에너지를 최소화한다.
    STORM 기반 world model에서는 과거 시계열을 요약하는 결정론적 hidden state $h_t$와 현재 관찰로부터 추론되는 이산 잠재 상태 $s_t$를 사용하므로, 단일 시점의 변분 자유 에너지는 다음과 같이 쓸 수 있다:
}
\EN{
    Under the free energy principle, the agent minimizes variational free energy to explain observations and expected free energy to select actions.
    In the STORM-based world model, temporal information is summarized by a deterministic hidden state $h_t$, while the current observation is encoded into a discrete latent state $s_t$, so the single-timestep variational free energy can be written as:
}

\begin{equation*}
    \begin{aligned}
        \mathcal{F} &= \mathbb{E}_{q(s_{t})}[\log q(s_{t}) - \log p(o_{t},s_{t})] \\
        &= \kldiv{q(s_t)}{p(s_t)} - \mathbb{E}_{q(s_t)}[\log p(o_t \mid s_t)] \\
        &= \kldiv{q_\phi(s_t \mid o_t)}{p_\phi(\hat{s}_t \mid h_t)} \\
        &\phantom{....}- \mathbb{E}_{q_\phi(s_t \mid o_t)}[\log p_\phi(\hat{o}_t \mid s_t)]
    \end{aligned}
\end{equation*}

\KO{
    정책 네트워크와 가치 네트워크를 학습시키는데 쓰일 기대 자유 에너지는 다음과 같다:
}
\EN{
    The expected free energy used to train the policy and value networks is given by:
}

\begin{equation*}
    \begin{aligned}
    \mathcal{G} &= \mathbb{E}_{q(o_t, s_t, a_t)}[\log q(s_t, a_t) - \log \tilde{p}(o_t, s_t, a_t)] \\
    &= \mathbb{E}_{q(o_t, s_t, a_t)}[\log q(a_t \mid s_t) + \log q(s_t \mid s_{t-1}, a_{t-1}) \\ 
    & \phantom{.....................} - \log p(a_t) - \log p(o_t \mid s_t) - \log \tilde{p}(s_t)]
    \end{aligned}
\end{equation*}

\KO{
    여기서 행동 사전 확률 $p(a_t)$는 균등 분포로 가정되므로 $\mathbb{E}_{q(o_t,s_t,a_t)}[-\log p(a_t)]$는 상수가 된다. 또한, 관찰 모호성 항 $\mathbb{E}_{q(o_t,s_t,a_t)}[-\log p(o_t \mid s_t)]$은 관찰 디코더가 고정된 분산을 갖는 가우시안 출력을 사용하므로 상태 $s_t$에 의존하지 않는 상수가 되어 소거할 수 있다. 나아가 본 논문에서는 self-prior에 의한 행동 생성 메커니즘에 집중하기 위해, 잠재 상태의 전이 불확실성과 관련된 항 $\mathbb{E}_{q(o_t,s_t,a_t)}[\log q(s_t \mid s_{t-1}, a_{t-1})]$을 상수로 취급하여 생략하였다. 최종적으로 사용된 기대 자유 에너지는 다음과 같다:
}
\EN{
    Here, the action prior $p(a_t)$ is assumed to be uniform, so $\mathbb{E}_{q(o_t,s_t,a_t)}[-\log p(a_t)]$ becomes a constant.
    In addition, the observation ambiguity term $\mathbb{E}_{q(o_t,s_t,a_t)}[-\log p(o_t \mid s_t)]$ can be eliminated because the observation decoder uses a Gaussian output with fixed variance, making it a constant independent of the state $s_t$.
    Furthermore, to focus on the action-generation mechanism driven by the self-prior, we treat the term $\mathbb{E}_{q(o_t,s_t,a_t)}[\log q(s_t \mid s_{t-1}, a_{t-1})]$ associated with the transition uncertainty of the latent state as a constant and omit it.
    The expected free energy ultimately used is given by:
}

\begin{equation*}
    \begin{aligned}
    \mathcal{G} &\approx - \mathbb{E}_{q_\phi(s_t)}[\log \tilde{p}_\xi(s_t) + \mathcal{H}(q(a_t \mid s_t))]
    \end{aligned}
\end{equation*}

\KO{
    단, 행동 엔트로피 항은 정책 학습시에 $\mathcal{H}(\pi_\theta)$로서 별도로 계산한다. 따라서 self-prior가 스티커가 없는 상태에서 학습되었다면, 스티커가 부착된 관찰에 대응하는 잠재 상태는 낮은 확률을 갖게 되며, 자유 에너지를 낮추기 위해 에이전트는 익숙한 상태로 복귀하는 방향의 행동을 선택하게 된다.
}
\EN{
    Note that the action entropy term is computed separately as $\mathcal{H}(\pi_\theta)$ during policy training.
    Therefore, if the self-prior is trained on sticker-free states, latent states corresponding to sticker-bearing observations receive low probability, and the agent selects actions that return it to familiar states so as to reduce free energy.
}

\subsection{Training}

\KO{
    에이전트의 학습은 세 단계에 걸쳐 점진적으로 활성화된다.
    먼저 world model을 학습하여 잠재 표현을 안정화한 뒤 그 위에서 self-prior를, 다시 그 위에서 정책을 학습한다. 이는 각 모듈이 안정된 기반 위에서 학습되도록 하기 위해 고안되었으나, 아래의 임계값(에피소드 100/120/140)은 총 50,000 에피소드에 이르는 전체 학습의 초반 일부에 불과하며 수렴 시점의 결과에 미치는 영향은 미미할 것으로 예상된다. 수집과 학습은 10 에피소드마다 교대되었으며, 한 번 학습시 100 train step이 진행되었다.
}
\EN{
    Training is activated progressively in three stages.
    We first train the world model to stabilize the latent representation, then train the self-prior on top of it, and finally train the policy on top of the self-prior. This scheme is designed so that each module is trained on a stable foundation; however, the thresholds below (episodes 100/120/140) constitute only the initial portion of the entire training of 50,000 episodes, so their effect on the results at convergence is expected to be negligible. Data collection and training were alternated every 10 episodes, and each training session consisted of 100 training steps.
}

\subsubsection{World Model Training}

\KO{
    World model은 에피소드 100 이후부터 학습을 시작한다.
    손실 함수는 재구성 손실, 보조 decoder 손실, 그리고 KL 발산 손실의 합으로 구성된다:
}
\EN{
    World model training begins after episode 100.
    The loss function combines reconstruction loss, an auxiliary decoder loss, and KL divergence losses:
}

\begin{equation*}
    \mathcal{L}_{\phi} = \mathcal{L}_{recon} + \mathcal{L}_{dec} + \beta_{dyn} \mathcal{L}_{dyn} + \beta_{rep} \mathcal{L}_{rep}
\end{equation*}

\KO{
    재구성 손실 $\mathcal{L}_{recon}$은 posterior latent로부터 복원한 시각 및 고유수용감각 입력에 대한 MSE 손실이다.
    보조 decoder 손실 $\mathcal{L}_{dec}$은 디코더 경로를 따라 계산되는 보조 재구성 항으로, 잠재 상태에 stop-gradient를 적용하여 인코더와 posterior, 역믹서는 갱신하지 않고 decoder만 갱신한다. 이 손실은 디버깅을 위해 사용되었고, $\mathcal{L}_{recon}$만으로도 학습이 가능하므로 선택적 항이며, stop-gradient 덕분에 world model의 주 목적 함수에는 영향을 주지 않는다.
    동역학 손실 $\mathcal{L}_{dyn}$과 표현 손실 $\mathcal{L}_{rep}$은 prior와 posterior 간의 KL 발산으로, stop-gradient와 free bits를 적용한다:
}
\EN{
    The reconstruction loss $\mathcal{L}_{recon}$ is the MSE loss for visual and proprioceptive inputs reconstructed from the posterior latent.
    The auxiliary decoder loss $\mathcal{L}_{dec}$ is an auxiliary reconstruction term computed along the decoder path; by applying a stop-gradient to the latent state, it updates only the decoder while leaving the encoder, posterior, and inverse mixer unchanged. This loss was used for debugging and is optional, since training is feasible with $\mathcal{L}_{recon}$ alone, and thanks to the stop-gradient it does not affect the main objective of the world model.
    The dynamics loss $\mathcal{L}_{dyn}$ and representation loss $\mathcal{L}_{rep}$ are KL divergences between prior and posterior, with stop-gradient and free bits:
}

\begin{equation*}
    \begin{aligned}
    \mathcal{L}_{\text{dyn}} &= \max\big(1,\, \kldiv{\operatorname{sg}(q_\phi(s_{t} \mid o_{t}))}{ p_\phi(\hat{s}_{t} \mid h_t)}\big) \\
    \mathcal{L}_{\text{rep}} &= \max\big(1,\, \kldiv{q_\phi(s_{t} \mid o_{t})}{\operatorname{sg}(p_\phi(\hat{s}_{t} \mid h_t))}\big)
    \end{aligned}
\end{equation*}

\KO{
    여기서 $\operatorname{sg}(\cdot)$는 stop-gradient 연산자이며, $\max(1, \cdot)$는 free bits로 posterior collapse를 방지한다.
}
\EN{
    Here $\operatorname{sg}(\cdot)$ is the stop-gradient operator and $\max(1, \cdot)$ applies free bits to prevent posterior collapse.
}

\subsubsection{Self-Prior Training}

\KO{
    Self-prior는 에피소드 120 이후부터 학습을 시작한다.
    손실 함수는 autoregressive cross-entropy와 slow target에 대한 KL 정규화로 구성된다:
}
\EN{
    Self-prior training begins after episode 120.
    The loss function comprises autoregressive cross-entropy and KL regularization against a slow target:
}

\begin{equation*}
    \mathcal{L}_{\xi} = -\frac{1}{K}\sum_{k=1}^{K} \log \tilde{p}_\xi(s^{(k)} \mid s^{(<k)}) + \kldiv{\tilde{p}^{\text{EMA}}_{\xi}}{\tilde{p}_\xi}
\end{equation*}

\KO{
    여기서 $\tilde{p}^{\text{EMA}}_{\xi}$는 EMA 방식으로 업데이트되는 target 네트워크이다.
}
\EN{
    Here $\tilde{p}^{\text{EMA}}_{\xi}$ is a target network updated via exponential moving average (EMA).
}

\subsubsection{Policy Training}

\KO{
    정책 학습은 에피소드 140 이후부터 시작되며, world model 내부에서의 상상(imagination)을 통해 이루어진다.
    실제 관찰로부터 초기 hidden state를 구성한 뒤, 현재 정책을 따라 $H=16$ 스텝의 미래 궤적을 상상한다.
    각 상상된 시점에서 self-prior가 기대 자유 에너지를 계산하고, GAE~\cite{schulman_high-dimensional_2015}로 $\lambda$-return을 산출한다.
    정책과 가치 네트워크의 손실은 다음과 같다:
}
\EN{
    Policy training begins after episode 140 and is performed through imagination within the world model.
    An initial hidden state is built from real observations, and future trajectories of $H=16$ steps are imagined under the current policy.
    At each imagined step the self-prior computes expected free energy, and $\lambda$-returns are computed via GAE~\cite{schulman_high-dimensional_2015}.
    The policy and value networks are trained with the following losses:
}

\begin{equation*}
    \begin{aligned}
\mathcal{L}_{\theta}^{\pi} &= \sum\nolimits_t \Big[-\operatorname{sg}\Big(\frac{G^\lambda_t - V_{\psi}(s_t, h_t)}{\max(1, S)} \Big) \ln \pi_{\theta}(a_t \mid s_t, h_t) \\
&\phantom{.............................................} - \eta \mathcal{H}(\pi_\theta(a_t \mid s_t, h_t)) \Big] \\
\mathcal{L}_{\psi} &= \sum\nolimits_t \Big[ \Big(V_\psi(s_t, h_t) - \operatorname{sg}(G^\lambda_t) \Big)^2 \\
&\phantom{..........} + \Big( V_\psi(s_t, h_t) - \operatorname{sg}(V^{\text{EMA}}_{\psi}(s_t, h_t))\Big)^2 \Big] \\
G^\lambda_t &= \mathcal{G}(s_t) \\
&\phantom{....} + \gamma 
    \begin{cases}
      (1 - \lambda) V_\psi(s_{t+1}, h_{t+1}) + \lambda G^\lambda_{t+1}, & \text{if}\ t<H \\
      V_\psi(s_H, h_H), & \text{if}\ t=H
    \end{cases}
    \end{aligned}
\end{equation*}

\KO{
    여기서 $S$는 $\lambda$-return의 정규화 항이다.
    World model과 self-prior는 AdamW를, 정책과 가치 네트워크는 Adam을 사용하며, 그래디언트 클리핑은 ZClip~\cite{kumar2025zclip}으로 수행한다.
}
\EN{
    Here $S$ is a normalization term for $\lambda$-returns.
    The world model and self-prior use AdamW, while the policy and value networks use Adam; gradient clipping is performed with ZClip~\cite{kumar2025zclip}.
}

\section{EXPERIMENTS AND RESULTS}

\KO{
    우리의 실험은 스티커가 없는 일상적 상황에서 self-prior를 학습한 에이전트가, 거울 앞에서 자기 몸에 새롭게 부착된 스티커를 스스로 찾아 제거할 수 있는지를 평가한다.
    여기서 한 에피소드는 에이전트의 자세를 초기화한 상태에서 시작하여 고정된 수의 제어 스텝 동안 진행되는 하나의 롤아웃을 의미하며, 스티커가 포함되는 에피소드에서는 시작 시점에 스티커가 배치된다.
    학습 전체에서는 총 50,000 에피소드 분량의 경험을 수집하였다 (총 학습 단계 500,000).
    각 에피소드의 행동은 무작위 정책과 학습된 정책이 절반씩 혼합되어 생성되었으며, 스티커가 부착된 에피소드와 부착되지 않은 에피소드 역시 절반씩 수집되었다.
    평가 에피소드 시작 시 스티커는 머리 또는 상반신 몸통 표면 가운데 전방 \qty{60}{\degree} 이내를 향하는 위치에 무작위로 배치된다.

    변분 자유 에너지 최소화를 위한 world model 학습과, 기대 자유 에너지 최소화를 위한 정책 학습에서는 스티커의 존재 여부나 행동 방식에 관계없이 모든 데이터를 사용하여, 다양한 상황에서 올바른 예측과 행동 선택을 학습하도록 하였다.
    반면 self-prior 학습에는 수집된 데이터 중 스티커가 부착된 에피소드를 약 5\%만 포함하도록 별도로 추출하여, 에이전트가 ``스티커가 붙어 있지 않은 나''라고 하는 일상적인 경험을 모사하였다.
    다시 강조하면, 에이전트에게 스티커를 떼라고 명시적으로 지시한 것이 아니다.

    스티커 제거는 에이전트의 손이 스티커로부터 \qty{2}{\cm} 이내에 연속 $50$ 시뮬레이션 스텝 (0.5초) 동안 머물 때 성공으로 판정하였다.
    주요 지표는 스티커 제거 전후의 기대 자유 에너지, 에피소드 전체에 걸친 손-스티커 거리, 그리고 스티커 제거 확률이다.
}
\EN{
    Our experiments evaluate whether an agent whose self-prior was learned from everyday sticker-free situations can detect and remove a newly attached sticker on its own body while seated in front of a mirror.
    Here, an episode denotes a single rollout that starts from a reset of the agent's posture and proceeds for a fixed number of control steps; in sticker-bearing episodes, the sticker is placed at the start.
    The full learning process collected a total of 50,000 training episodes (500k training steps in total).
    Actions in each episode were generated from an equal mix of random and learned policies, and episodes with and without a sticker were collected in equal proportion.
    At the beginning of each evaluation episode, a sticker is placed randomly on a surface on the head or upper torso facing within \qty{60}{\degree} of the forward direction.

    For world model training (variational free energy minimization) and policy training (expected free energy minimization), all data were used regardless of sticker presence or action source, so that the models learn correct predictions and action selection across diverse situations.
    In contrast, self-prior training used data in which only about 5\% of episodes contained a sticker, causing the agent to form a prior of a sticker-free self.
    Crucially, the agent was never explicitly instructed to remove the sticker.

    Sticker removal was counted as successful when the agent's hand stayed within \qty{2}{\cm} of the sticker for 50 consecutive simulation steps (0.5 seconds).
    The main metrics are expected free energy before and after sticker removal, hand-sticker distance over the episode, and sticker-removal probability.
}

\subsection{Expected Free Energy as Mismatch Evidence}

\begin{figure}[t]
    \centering
    \subfloat[]{\includegraphics[width=0.45\textwidth]{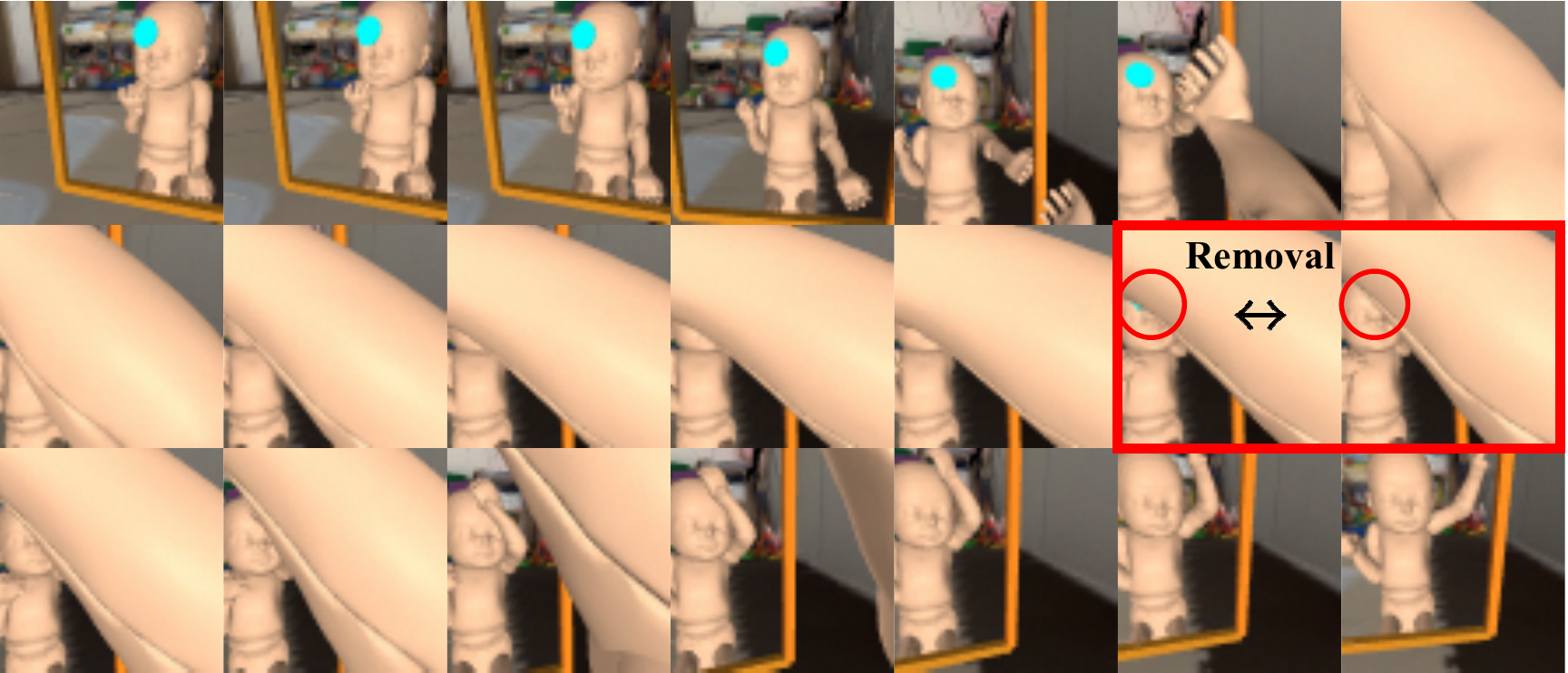}\label{fig:fig3a}}
    \hfil
    \subfloat[]{\includegraphics[width=0.49\columnwidth]{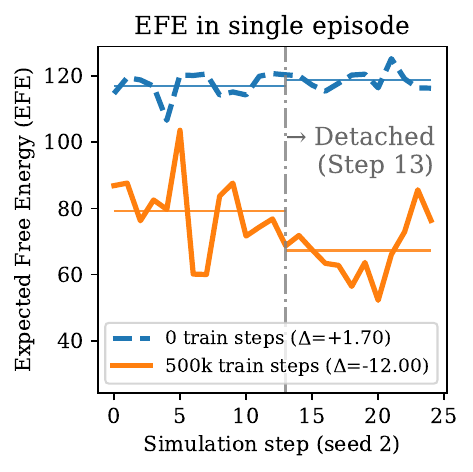}\label{fig:fig3b}}
    \hfil
    \subfloat[]{\includegraphics[width=0.49\columnwidth]{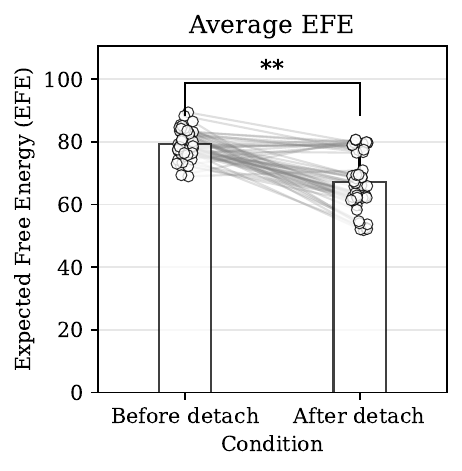}\label{fig:fig3c}}
    \caption{
    \KO{거울 검사에서의 스티커 제거 행동 창발 (seed 2). (a) Self-prior 학습이 진행된 에이전트가 거울 속 스티커를 발견하고 손을 뻗어 스티커를 제거하는 모습. 빨간색 강조된 부분이 스티커가 제거되는 순간이다. (b) 동일한 에피소드에서 모델 학습 전후의 기대 자유 에너지 비교. Self-prior가 학습되기 전에는 장면별 차이가 작지만, 학습 이후에는 스티커가 제거된 장면의 기대 자유 에너지가 더 낮다. (c) 전체 seed에서 스티커 제거 전후의 기대 자유 에너지 변화. 제거 이후 기대 자유 에너지가 낮아지는 경향을 보인다.}
    \EN{Emergence of sticker-removal behavior in the mirror test (seed 2).
    (a) An agent undergoing self-prior learning detects the sticker in the mirror, reaches out, and removes it.
    The red-highlighted region marks the exact moment of sticker removal.
    (b) Comparison of expected free energy before and after model learning in the same episode.
    Before self-prior learning, scene-wise differences are small, whereas after learning the scene with sticker removal shows lower expected free energy.
    (c) Change in expected free energy before and after sticker removal across all seeds.
    Expected free energy tends to decrease after removal.}
    }
    \label{fig:fig3}
\end{figure}

\KO{
    스티커가 없는 상황에서 self-prior를 학습한 뒤 스티커를 부착하면, 에이전트는 자발적으로 스티커를 향해 손을 뻗는 행동을 보였다 (Fig.~\ref{fig:fig3a}).
    에이전트는 촉각 입력 없이 오직 시각과 고유수용감각만을 사용하여 거울 속 자신의 얼굴에 붙은 스티커의 위치를 파악하고 손을 정확하게 이동시켰다.
    이는 외부 보상이나 명시적 표식 위치 추정 모듈 없이도, 익숙한 자기 상태와 현재 관찰의 불일치를 줄이는 방향의 행동이 창발할 수 있음을 보여준다.
}
\EN{
    After the self-prior was trained on sticker-free situations, attaching a sticker induced spontaneous reaching toward the sticker (Fig.~\ref{fig:fig3a}).
    Using only vision and proprioception, without tactile input, the agent localized the sticker on its own face in the mirror and moved its hand accurately to it.
    This shows that behavior directed toward the mark can emerge without external reward or an explicit module for estimating sticker location, simply by reducing the mismatch between the current observation and a familiar self state.
}

\KO{
    Fig.~\ref{fig:fig3b}는 같은 행동 시퀀스에 대해 학습 전후의 기대 자유 에너지를 비교한 결과이다.
    Self-prior가 충분히 학습된 뒤에는 스티커가 부착된 상태에서 기대 자유 에너지가 높고, 스티커가 제거된 이후에는 기대 자유 에너지가 낮아진다.     예시 에피소드에서 제거 이후의 평균 기대 자유 에너지는 제거 이전보다 $12.00$ 낮았다.
    반대로 학습 전 모델에서는 제거 전후의 차이가 뚜렷하게 나타나지 않는다.

    전체 80회 평가에서도 스티커 제거 후의 기대 자유 에너지($67.33 \pm 8.94$)는 제거 전($79.33 \pm 4.34$)보다 대체적으로 낮았으며, 이 차이는 Wilcoxon signed-rank test에서 유의하였다 ($p = 6.33 \times 10^{-9}$; Fig.~\ref{fig:fig3c}).
    이는 학습된 self-prior가 현재 장면을 ``익숙한 자기 상태''와 비교하는 내부 기준으로 작동하고 있음을 뒷받침한다.
}
\EN{
    Fig.~\ref{fig:fig3b} compares expected free energy along the same action sequence before and after learning.
    Once the self-prior is sufficiently learned, expected free energy is high while the sticker is present and decreases after the sticker is removed. In the illustrative episode, the mean expected free energy after removal was $12.00$ lower than before removal.
    By contrast, the untrained model shows no clear difference before and after removal.

    Across all 80 evaluations, expected free energy after sticker removal ($67.33 \pm 8.94$) was generally lower than before removal ($79.33 \pm 4.34$), and this difference was significant in a Wilcoxon signed-rank test ($p = 6.33 \times 10^{-9}$; Fig.~\ref{fig:fig3c}).
    This supports the interpretation that the learned self-prior functions as an internal criterion that compares the current scene with a familiar self state.
}

\subsection{Quantitative Learning Dynamics}

\begin{figure}[t]
    \centering
    \subfloat[]{\includegraphics[width=0.49\columnwidth]{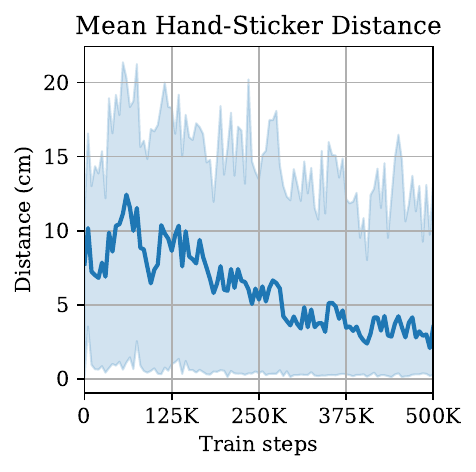}\label{fig:fig4a}}
    \hfil
    \subfloat[]{\includegraphics[width=0.49\columnwidth]{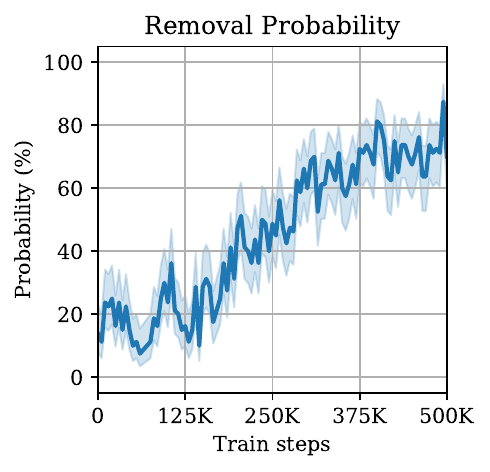}\label{fig:fig4b}}
    \caption{
    \KO{거울 검사의 정량적 결과. 실선은 모든 평가 에피소드의 평균값을 나타낸다. (a) 학습 과정에서 스티커와 손 사이 거리의 평균. 음영 밴드는 경험적 5--95 분위수 범위를 나타낸다. 스티커와의 거리는 양손 중 더 가까운 손을 기준으로 계산하였다. (b) 학습 과정에서의 스티커 제거 성공률 변화. 음영 밴드는 removal probability에 대한 Wilson score 95\% 신뢰구간을 나타낸다.}
    \EN{Quantitative results for the mirror test. Solid lines denote mean values aggregated over all evaluation episodes. (a) Mean hand-sticker distance during training. The shaded band indicates the empirical 5th--95th percentile range. Distance is computed using the hand that is closer to the sticker. (b) Sticker-removal success probability during training. The shaded band indicates the 95\% Wilson score confidence interval of the removal probability.}
    }
    \label{fig:quantitative_results}
\end{figure}

\KO{
    학습이 진행됨에 따라 에피소드 전체에 걸친 손과 스티커 사이 거리의 평균은 점진적으로 감소하였다 (Fig.~\ref{fig:fig4a}).
    주목할 점은, 스티커 제거 성공률이 아직 절반에 미치지 못하는 구간에서도 스티커와의 평균 거리가 꾸준히 감소했다는 점이다.
    이는 본격적인 리칭 행동이 안정적으로 나타나기 전부터 에이전트가 거울 속 스티커를 이미 인식하고 그 위치로 접근하려 했을 가능성을 시사한다.

    스티커 제거 성공률은 학습 초기 단계에서는 약 20\% 수준에 머물렀으나, 학습이 진행되면서 최종적으로 약 70\%까지 상승하였다 (Fig.~\ref{fig:fig4b}).
    따라서 제안한 메커니즘은 표식을 향한 접근 행동뿐 아니라 실제 표식 제거 행동도 일정 수준까지 학습할 수 있음을 보여준다.

    정성적 관찰에 따르면 실패 에피소드에서는 손이 거울 속 스티커를 가리거나, 고개 회전으로 인해 스티커가 시야에서 사라지는 경우와 동반되었다.
    또한 일부 실패 에피소드의 스티커 위치에 대해 시뮬레이션상의 팔을 직접 조작해 본 결과, 흉부 측면과 같이 신체의 운동 범위로 인해 실제로 도달하기 어려운 위치가 있음을 확인하였다.
    이는 제거 성공률의 상한이 지각적 가시성뿐 아니라 운동학적 도달 가능성에도 영향을 받음을 시사한다.
}
\EN{
    As training progressed, the mean hand-sticker distance over the episode decreased gradually (Fig.~\ref{fig:fig4a}).
    Notably, even while the sticker-removal success rate remained below 50\%, the mean distance to the sticker continued to decrease steadily.
    This suggests that, before reliable reaching behavior was fully established, the agent may already have recognized the sticker in the mirror and attempted to move toward its location.

    The sticker-removal success rate stayed near 20\% in the early phase of training and increased to about 70\% by the end of training (Fig.~\ref{fig:fig4b}).
    Thus the proposed mechanism learns not only approach behavior toward the mark but also actual removal behavior to a meaningful extent.

    Qualitative inspection showed that failure episodes were accompanied by the hand occluding the sticker in the mirror or by head rotation causing the sticker to leave the visual field.
    In addition, by manually controlling the simulated arms for a subset of failed episodes, we confirmed that some sticker locations, such as the lateral sides of the torso, were physically difficult to reach because of limits in the body's range of motion.
    This suggests that the ceiling on removal success depends on both perceptual visibility and kinematic reachability.
}

\subsection{What the Self-Prior Has Learned}

\begin{figure}[t]
    \centering
    \subfloat[]{\includegraphics[width=0.43\textwidth]{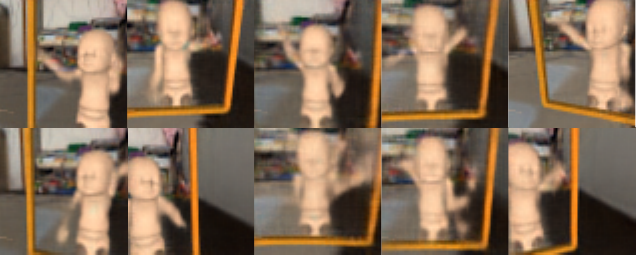}\label{fig:fig5a}}
    \par\vspace{1.5em}
    \subfloat[]{\includegraphics[width=0.45\textwidth]{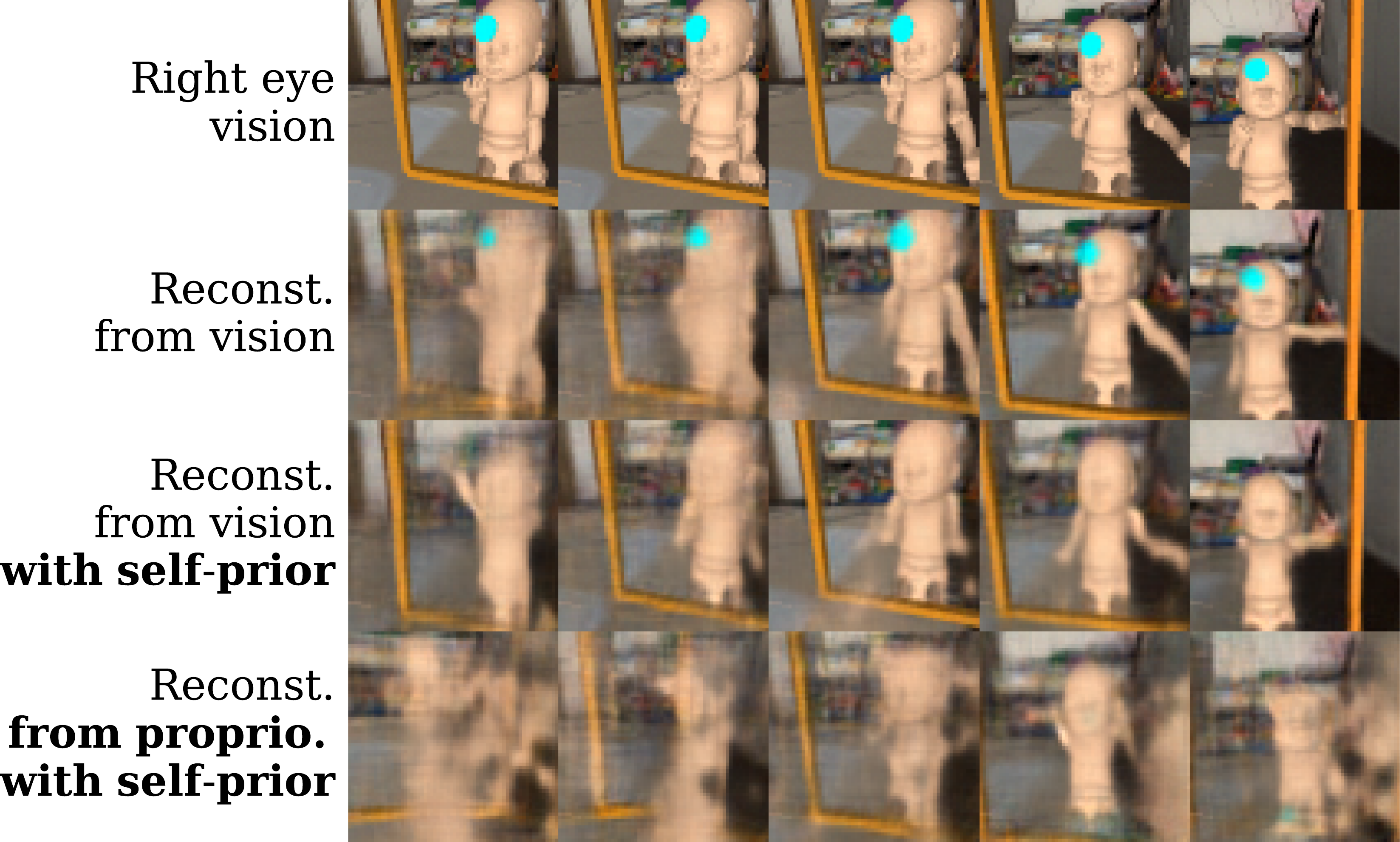}\label{fig:fig5b}}
    \caption{
    \KO{자신의 다감각 경험에 대한 밀도 모델인 Self-prior. (a) Self-prior로부터 무작위로 생성된 다감각 경험의 샘플들. 샘플들은 self-prior가 스티커가 없는 자신을 주로 학습했음을 보여준다. (b) 스티커가 있는 장면의 잠재변수를 self-prior에 통과시켜 재구성하면 스티커가 사라지며, 이는 self-prior가 스티커가 없는 신체 상태를 선호하는 분포임을 보여준다. 또한 고유수용감각만으로도 (시각 입력을 0으로 둔 채) 시각적 자기 모습을 복원할 수 있어, 감각 간 교차 모달 연관이 포착되었음을 확인할 수 있다.}
    \EN{Self-prior as a density model of the agent's multisensory self-experience. (a) Samples generated from the self-prior. The samples show that the self-prior predominantly captures a sticker-free self. (b) Reconstructing a sticker-bearing latent through the self-prior removes the sticker, indicating that the distribution favors a sticker-free body state. Furthermore, a visual self-appearance can be recovered from proprioception alone (with the visual input set to zero), confirming that cross-modal associations have been captured.}
    }
    \label{fig:self_prior}
\end{figure}

\KO{
    Fig.~\ref{fig:self_prior}는 self-prior가 익숙한 자기 경험의 구조를 실제로 학습했음을 보여준다.
    Self-prior는 잠재 변수에 대한 autoregressive 결합 분포이므로, BOS 토큰 $s_t^{(0)}$에서 시작해 $s_t^{(1)}, \ldots, s_t^{(32)}$를 차례로 샘플링하면 감각 입력 없이도 학습된 분포로부터 잠재 상태를 생성할 수 있다.
    이렇게 샘플링한 잠재 상태를 디코더로 복원한 결과는 (다양한 자세를 가진) 스티커가 없는 자신에 해당하며, 이는 self-prior가 sticker-free self를 선호하는 분포가 되었음을 시사한다 (Fig.~\ref{fig:fig5a}).
    
    또한 스티커가 있는 장면을 압축한 잠재변수에 대해 self-prior를 통과시킨 후 재구성(reconstruct)한 결과로부터, self-prior의 특징을 다시 한 번 확인할 수 있다.
    교차 모달 복원에서는 인코더에 시각 입력을 0으로 넣어 고유수용감각만으로 통합 잠재 상태를 구성한 뒤 시각을 복원하였다.
    복원된 시각 자기 모습은 다소 퍼져 있어 세밀한 형태보다는 자세 수준의 거친 공간 정보를 담고 있으나, 고유수용감각만으로 시각적 자기 모습을 복원할 수 있다는 점은 self-prior가 시각과 고유수용감각 사이의 연관 정보도 학습했음을 보여준다 (Fig.~\ref{fig:fig5b}).
}
\EN{
    Fig.~\ref{fig:self_prior} shows that the self-prior actually learns the structure of familiar self-experience.
    Because the self-prior is an autoregressive joint distribution over the latent variables, starting from the BOS token $s_t^{(0)}$ and sampling $s_t^{(1)}, \ldots, s_t^{(32)}$ in turn generates latent states from the learned distribution without any sensory input.
    Decoding such sampled latent states yields a sticker-free self in diverse poses, indicating that the distribution has indeed come to prefer a sticker-free body state (Fig.~\ref{fig:fig5a}).
    
    Reconstructing a sticker-bearing latent variable after passing it through the self-prior further confirms these characteristics.
    For the cross-modal reconstruction, we set the visual input to the encoder to zero and formed the unified latent state from proprioception alone before reconstructing vision.
    The recovered visual self-appearance is somewhat diffuse and conveys coarse, posture-level spatial information rather than fine detail; nonetheless, the ability to recover a visual self-appearance from proprioception alone shows that the self-prior has learned the association between vision and proprioception (Fig.~\ref{fig:fig5b}).
}

\section{DISCUSSION}

\KO{
    앞서 우리는 self-prior와 기대 자유 에너지 최소화라는 단일 원리만으로, 거울 속 표식을 향한 행동이 외부 보상 없이 창발할 수 있음을 확인하였다.
    에이전트는 스티커를 약 70\%의 확률로 제거하였으며, 제거 전후의 기대 자유 에너지 변화는 self-prior가 익숙한 자기 관련 상태를 선호하는 내부 기준으로 작동한다는 해석과 부합하였다.
    이하에서는 이러한 결과를 기존의 신체 표상 이론 및 거울 자기 인식 모델의 맥락에서 해석하고, 본 접근법의 한계와 향후 연구 방향을 논의한다.

    본 연구의 결과는 self-prior가 body schema와 기능적으로 유사한 역할을 할 수 있음을 시사한다.
    Body schema는 행동 계획과 운동 제어를 위한 감각운동적 신체 표현으로, 예를 들면 시각적 안내 없이도 자신의 신체 부위에 도달할 수 있게 하는 것과 같은 암묵적 지식에 해당한다~\cite{paillard1999body, gallagher_body_1986}.
    Fig.~\ref{fig:self_prior}에서 보이듯이 self-prior는 시각과 고유수용감각의 교차 모달 연관을 포착하며, 거울 과제에서는 시각 정보로부터 표식의 위치를 명시적으로 추출하지 않고도 다감각 불일치를 줄이는 방향으로 행동을 계획한다.
    이처럼 self-prior가 행동 계획을 직접 안내한다는 점에서 body schema 해석과 잘 부합한다.

    동기 측면에서도 self-prior는 기존 모델과 다른 의미를 갖는다.
    기존의 거울 자기 인식 계산론적 모델들은 시각적 이상의 감지와 행동 생성 사이에 명시적 매핑을 두거나~\cite{hoffmann_robot_mirror_2021}, 자타 구별에 초점을 맞추면서도~\cite{lanillos_robot_self_other_2020} 표식을 향한 행동의 동기 자체를 설명하지 않았다.
    본 연구에서는 self-prior가 이 동기를 자연스럽게 제공한다: 스티커 제거라는 목표가 사전에 설정된 것이 아니라, 현재 상태와 익숙한 자기 상태 사이의 불일치가 행동을 촉발한다.
    이는 탐색을 통한 외부 보상 획득에 초점을 맞추는 기존 내발적 동기 연구와 차별화된다.
    결과적으로 우리의 목적 함수는 내재적 보상을 사용하는 강화학습과 수식적으로 유사하지만, 그 보상이 환경에서 주어지는 것이 아니라 에이전트가 학습한 self-prior 하에서 스스로 평가하는 로그 밀도라는 점에서 개념적으로 구별된다.

    더 나아가, 선호 분포를 수작업으로 지정하는 표준 능동 추론과 달리, 본 연구는 그 선호 분포 자체를 에이전트의 경험으로부터 학습한다.
    SR-AIF~\cite{nguyen2025sr} 또한 학습된 동적 상태 선호를 사용한다는 점에서 본 연구와 관련된다.
    다만 그들의 선호는 희소 보상 목표 과제를 풀기 위한 목표 지향적 선호인 반면, 본 연구의 self-prior는 과제나 목표 없이 에이전트 자신의 일상적 다감각 경험의 밀도를 모델링하여 발달적 창발을 탐구한다는 점에서 차이가 있다.

    Zaadnoordijk와 Bayne~\cite{zaadnoordijk_origins_2020}은 의도적 주체성의 기원에 관해 자극 유도 의도(stimulus-elicited intention)와 내재적 의도(endogenous intention)를 구별하였는데, self-prior에 의해 유발되는 스티커 제거 행동은 전자에 해당한다.
    이러한 메커니즘은 유아가 자극 유도 의도에 의해 행동하다가 점차 내재적 의도를 형성해 나간다는 발달론적 관점과 일치한다.

    본 연구의 한계로서, 기존의 연구들은 실제 로봇에서 검증된 반면~\cite{hoffmann_robot_mirror_2021, lanillos_robot_self_other_2020}, 본 연구는 시뮬레이션 환경에 한정된다는 한계가 있다.
    제거 확률이 약 70\% 수준에 머문 점도 한계인데, 앞서 보고한 실패 사례들은 표식을 물리적으로 제거하지 않더라도 시각적으로 보이지 않게 만드는 것만으로 self-prior와의 불일치를 부분적으로 줄일 수 있음을 시사한다.
    이는 본 모델이 거울 자기 인식의 핵심 행동을 재현하더라도, 그 행동의 의미 부여는 여전히 환경 구조와 신체 역학의 영향을 강하게 받는다는 뜻이기도 하다.

    그렇다면 본 모델은 거울 자기 인식을 어느 수준까지 구현하는가?
    Apps와 Tsakiris~\cite{apps_free-energy_2014}는 자기 인식이 자유 에너지 원리 하에서 확률적으로 모델링될 수 있으며, 자신의 신체가 베이지안적으로 ``나''일 가능성이 가장 높은 것으로 추론된다고 제안하였다.
    Self-prior는 이러한 확률적 자기 표상의 계산론적 구현에 가깝다고 볼 수 있으며, 익숙한 감각 경험에 높은 밀도를 할당함으로써 익숙한 자기 관련 상태와 그렇지 않은 상태를 암묵적으로 구별한다.
    Rochat~\cite{rochat2003five}의 자기 인식 5단계 분류에서, 본 모델은 반사 이미지를 자신의 몸에 연결할 수 있는 수준 3: 식별(identification)에 가장 가깝다.

    한편, Mitchell~\cite{mitchell_mental_1993}의 거울 자기 인식에 관한 두 이론 중, 귀납적 이론은 (1) 성숙한 운동감각-시각 매칭과 (2) 거울 대응의 이해를 요구하며, 연역적 이론은 여기에 (3) 완전한 대상 영속성의 이해와 (4) 신체 부위의 대상화를 추가로 요구한다.
    본 모델은 self-prior를 통한 시각-고유수용감각 매칭과 거울 반사의 암묵적 학습을 통해 귀납적 이론의 핵심 요소를 구현하지만, 대상 영속성이나 신체 부위의 명시적 대상화는 구현되지 않았다.

    또한 mark-directed behavior가 곧바로 고차원적 자기 의식의 증거라는 보다 강한 주장과는 거리를 두어야 한다.
    표식을 향한 행동은 거울 자기 인식의 필요 조건이지만 충분 조건은 아니며, 본 연구는 그 필요한 행동이 어떻게 창발하는지를 보인 것이다.
    실제로 Kohda 등~\cite{kohda_cleaner_fish_2022}은 특정 어종이 생태학적으로 관련된 색상의 마크에 반응하여 마크 테스트를 통과한다고 보고하였으나, Gallup와 Anderson~\cite{gallup_self-recognition_2020}은 이러한 해석의 엄밀성에 의문을 제기하는 등, mark-directed behavior의 해석을 둘러싼 논쟁은 여전히 계속되고 있다.
    따라서 본 연구는 거울 자기 인식 전체를 완전히 설명했다고 주장하기보다, self-prior라는 간결한 메커니즘이 그 핵심 행동을 어떻게 생성할 수 있는지를 보여주는 계산론적 가설로 해석되어야 한다.

    이러한 이론적 논의에 더하여, 모델이 다루는 감각 양상과 환경 조건의 확장도 필요하다.
    현재 모델은 촉각 입력을 포함하지 않으나, Chinn 등~\cite{chinn_tactile_2020}은 신체 촉각 목표에 도달하는 경험을 제공받은 유아가 더 일찍 거울 자기 인식을 달성한다고 보고하였으며, 향후 촉각 양상의 통합이 학습 효율을 향상시킬 수 있다.
    또한 거울의 각도나 거리를 변화시킨 조건에서의 일반화 능력을 검증하고, 실제 로봇 및 영아 발달 데이터와의 비교를 통해 모델의 타당성을 확인하는 것이 향후 과제로 남는다.
}
\EN{
    We have demonstrated that, under the single principle of a self-prior combined with expected free energy minimization, mark-directed behavior emerges without external rewards.
    The agent removed stickers with approximately 70\% probability, and the change in expected free energy before and after removal was consistent with the self-prior acting as an internal criterion that favors familiar self-related states.
    In what follows, we interpret these findings in the context of existing theories of body representation and computational models of mirror self-recognition, and discuss the limitations and future directions of the present approach.

    The present results suggest that the self-prior can play a role functionally analogous to the body schema.
    The body schema is a sensorimotor representation of the body for action planning and motor control, corresponding to implicit knowledge such as the ability to reach one's own body parts without visual guidance~\cite{paillard1999body, gallagher_body_1986}.
    As shown in Fig.~\ref{fig:self_prior}, the self-prior captures cross-modal associations between vision and proprioception, and in the mirror task it plans action toward reducing multisensory mismatch without explicitly extracting the sticker location from visual input.
    In this way, the self-prior aligns well with a body-schema interpretation in that it directly guides action planning.

    The self-prior is also distinctive in motivational terms.
    Existing computational models of mirror self-recognition either introduce explicit mappings between visual anomaly detection and action generation~\cite{hoffmann_robot_mirror_2021} or focus on self-other distinction~\cite{lanillos_robot_self_other_2020} without explaining the source of motivation that drives behavior toward the mark.
    In the present model, the self-prior naturally provides this motivation: the agent has no predefined goal of sticker removal, and behavior is automatically elicited by the mismatch between the current state and the familiar self state.
    This is distinct from existing intrinsic motivation work that focuses on improving exploration for external reward acquisition.
    Consequently, although our objective is mathematically similar to reinforcement learning with an intrinsic reward, it differs conceptually in that the reward is not supplied by the environment but is the log-density that the agent itself evaluates under its learned self-prior.

    Furthermore, unlike standard active inference, in which the preferred distribution is specified by hand, our approach learns this preferred distribution itself from the agent's own experience.
    SR-AIF \cite{nguyen2025sr} is also related to the present work in that it uses learned dynamic state preferences.
    However, their preference is goal-directed and aimed at solving sparse-reward tasks, whereas our self-prior models the density of the agent's own everyday multisensory experience without any task or goal, in order to study developmental emergence.

    Zaadnoordijk and Bayne~\cite{zaadnoordijk_origins_2020} distinguished stimulus-elicited intention from endogenous intention in the origins of intentional agency; the sticker-removal behavior induced by the self-prior corresponds to the former.
    This mechanism aligns with the developmental view that infants initially act on stimulus-elicited intentions and gradually form endogenous intentions.

    As a limitation, whereas prior work has been validated on real robots~\cite{hoffmann_robot_mirror_2021, lanillos_robot_self_other_2020}, the present study is confined to a simulated environment.
    The removal probability plateauing at approximately 70\% is also a limitation; the failure modes reported above suggest that mismatch with the self-prior can be partially reduced merely by making the sticker visually disappear rather than physically removing it.
    This also implies that, even though the model reproduces the core behavior of mirror self-recognition, the significance attributed to that behavior is still strongly influenced by environmental structure and body mechanics.

    To what level, then, does this model implement mirror self-recognition?
    Apps and Tsakiris~\cite{apps_free-energy_2014} proposed that self-recognition can be modeled probabilistically under the free energy principle, with one's body inferred as the entity most likely to be ``me'' in a Bayesian sense.
    The self-prior can be viewed as a computational implementation of such a probabilistic self-representation, implicitly separating familiar self-related states from less familiar ones by assigning high density to familiar sensory experiences.
    In Rochat's~\cite{rochat2003five} five levels of self-awareness, the model is closest to Level 3: identification, in which the reflected image can be linked to one's own body.

    Of Mitchell's~\cite{mitchell_mental_1993} two theories of mirror self-recognition, the inductive theory requires (1) mature kinesthetic-visual matching and (2) understanding of mirror correspondence, while the deductive theory additionally requires (3) full understanding of object permanence and (4) objectification of body parts.
    The model implements the core elements of the inductive theory through self-prior-based visual-proprioceptive matching and implicit learning of mirror reflection, but object permanence and explicit objectification of body parts are not yet implemented.

    We also deliberately stop short of the stronger claim that mark-directed behavior by itself constitutes evidence of higher-order self-consciousness.
    Touching the marked body part is a necessary but not a sufficient condition for mirror self-recognition; we show how this necessary behavior can emerge, rather than claiming that the model fully recognizes its reflection as itself.
    Indeed, the interpretation of mark-directed behavior remains actively debated: Kohda et al.~\cite{kohda_cleaner_fish_2022} reported that a fish species passes the mark test by reacting to ecologically relevant colored marks, while Gallup and Anderson~\cite{gallup_self-recognition_2020} questioned the rigor of such interpretations.
    We therefore interpret the present study not as a complete account of mirror self-recognition, but as a computational hypothesis showing how a concise self-prior mechanism can generate its key behavior.

    In addition to these theoretical points, extending the sensory modalities and environmental conditions addressed by the model is needed.
    The current model does not include tactile input; Chinn et al.~\cite{chinn_tactile_2020} showed that infants who received experience reaching for tactile targets on the body achieved mirror self-recognition earlier, suggesting that integrating the tactile modality may improve learning efficiency.
    Validating generalization under varying mirror angles and distances and comparing with real robots and infant developmental data remain future work.
}

\section{CONCLUSIONS}

\KO{
    본 연구는 self-prior와 기대 자유 에너지 최소화를 결합하여, 외부 보상 없이 거울 속 스티커를 인식하고 제거하는 행동이 창발함을 보였다.
    Self-prior는 body schema와 기능적으로 유사한 확률적 자기 표상으로 볼 수 있으며, 이는 시각-고유수용감각 간의 교차 모달 연관성을 암묵적으로 학습하고, 익숙한 자기 상태와의 불일치로부터 자극 유도 의도에 의해 자동적으로 행동을 유발한다.
    이는 거울 자기 인식의 귀납적 이론~\cite{mitchell_mental_1993}이 요구하는 운동감각-시각 매칭과 거울 대응의 이해를 계산론적으로 구현한 것에 해당한다.

    기존의 거울 자기 인식 계산론적 모델들이 가공된 특징이나 명시적 좌표 변환에 의존하는 것과 달리, 본 연구는 원시 감각 입력으로부터 기대 자유 에너지 최소화라는 단일 원리로 이상 감지에서 행동 생성까지 통합적으로 수행한다.
    또한 동기의 원천이 별도로 설계되지 않아도 self-prior로부터 자연스럽게 제공된다는 점에서, 자유 에너지 원리가 자기 인식의 발달적 기원을 탐구하기 위한 간결한 가설로서의 가능성을 제시한다.

    향후 연구에서는 촉각 양상의 통합과 다양한 거울 조건에서의 일반화 검증을 통해 모델의 적용 범위를 넓히고, 실제 로봇 및 영아 발달 데이터와의 비교를 통해 타당성을 확인할 필요가 있다.
    나아가, 대상 영속성과 신체 부위의 대상화를 포함하는 연역적 이론의 요소를 모델에 통합함으로써, Rochat의 수준 4 이상의 자기 인식으로의 발달 경로를 탐구해야 할 것이다.
}
\EN{
    This study demonstrated that combining a self-prior with expected free energy minimization gives rise to spontaneous sticker-recognition and removal behavior without external rewards.
    The self-prior can serve as a probabilistic self-representation functionally analogous to the body schema, implicitly learning cross-modal associations between vision and proprioception and automatically eliciting behavior through stimulus-elicited intention when a mismatch with the familiar self state is detected.
    This constitutes a computational implementation of the inductive theory of mirror self-recognition~\cite{mitchell_mental_1993}, which requires kinesthetic-visual matching and understanding of mirror correspondence.

    Unlike existing computational models of mirror self-recognition that rely on processed features or explicit coordinate transformations, this approach unifies anomaly detection and action generation under the single principle of expected free energy minimization from raw sensory input.
    The source of motivation is naturally provided by the self-prior without separate design, suggesting the potential of the free energy principle as a concise hypothesis for investigating the developmental origins of self-awareness.

    Future work should extend the model by integrating tactile modality, testing generalization under varying mirror conditions, and validating it through comparison with real robots and infant developmental data.
    Incorporating elements of the deductive theory, such as object permanence and objectification of body parts, would enable exploration of developmental pathways toward Level 4 and beyond in Rochat's framework.
}

\addtolength{\textheight}{-8.7cm}   

\section*{APPENDIX}

\subsection{Implementation Details}

\begin{table}[ht]
    \centering
    \caption{Controllable joint ranges used by the agent}
    \label{tab:joint_ranges}
    \begin{tabular}{lll}
    \hline
    \textbf{Category} & \textbf{Joint} & \textbf{Range (rad)} \\
    \hline
    \multirow{2}{*}{Neck} & pitch & $[-0.30, 0.10]$ \\
                          & yaw   & $[-0.30, 0.30]$ \\
    \hline
    \multirow{3}{*}{Shoulder} & elv\_angle    & $[-0.87, 2.09]$ \\
                              & shoulder\_elv & $[\phantom{-}0.17, 2.70]$ \\
                              & shoulder\_rot & $[-0.87, 0.79]$ \\
    \hline
    \multirow{2}{*}{Elbow} & elbow\_flexion & $[\phantom{-}0.00, 2.27]$ \\
                           & pro\_sup       & $[-1.57, 1.57]$ \\
    \hline
    \end{tabular}
\end{table}

\begin{table}[ht]
    \centering
    \footnotesize
    \caption{Architecture details of the proposed model}
    \label{tab:arch_details}
    \begin{tabular}{p{0.33\columnwidth}p{0.57\columnwidth}}
    \hline
    \textbf{Component} & \textbf{Specification} \\
    \hline
    Visual encoder & 4 convolutional layers, $s_t^\modalfont{V} \in \mathbb{R}^{1024}$\\
    Proprioceptive encoder & 2-layer MLP (32 hidden units),\\
    &$s_t^\modalfont{P} \in \mathbb{R}^{1024}$\\
    Observation mixer & 1 linear layer, $\bar{s}_t \in \mathbb{R}^{1024}$\\
    Action mixer & 1 linear layer, $e_t \in \mathbb{R}^{512}$ \\
    Sequence model & GPT-like Transformer decoder, \\
    & 2 layers, 8 heads, dropout 0.1 \\
    Dynamics predictor & 1 linear layer \\
    Visual decoder & 1 linear layer and 4 convolutional layers, \\
    &Gaussian output with fixed std 1 \\
    Proprioceptive decoder & 3-layer MLP, \\
    &Gaussian output with fixed std 1 \\
    Policy network & 3-layer MLP (128 hidden units),\\
    &$[s_t; h_t] \in \mathbb{R}^{1536}$, \\
    & tanh-transformed normal output \\
    Value network & 3-layer MLP (256 hidden units),\\
    &$[s_t; h_t] \in \mathbb{R}^{1536}$, \\
    & symlog two-hot output (255 classes) \\
    \hline
    \end{tabular}
\end{table}

\begin{table}[ht]
    \centering
    \footnotesize
    \caption{Training and implementation details}
    \label{tab:train_impl_details}
    \begin{tabular}{p{0.40\columnwidth}p{0.50\columnwidth}}
    \hline
    \textbf{Setting} & \textbf{Value} \\
    \hline
    Hardware / software & Ubuntu 22.04.5 LTS, \\
    & Intel Xeon E5-2697 v4, \\
    & NVIDIA RTX A6000, \\
    & Python 3.13.7, PyTorch 2.7.1 \\
    Image resolution & $64 \times 64 \times 3$ RGB \\
    Simulation time step & \qty{0.01}{\s} \\
    Action repeat & 5 \\
    Episode length & 100 steps ($\qty{0.01}{\s} \times 5 \times 100 = \qty{5}{\s}$) \\
    Training schedule & 100 train steps per 10 episodes \\
    Data collection policy mix & Random \& policy mixed 50/50 \\
    Replay buffer size & 1000 episodes \\
    Evaluation frequency & Every 500 episodes \\
    Evaluation runs & 8 seeds, 10 runs per seed \\
    $\beta_{dyn}$ & 0.5 \\
    $\beta_{rep}$ & 0.1 \\
    $\lambda$ & 0.95 \\
    $\gamma$ & 0.997 \\
    EMA decay & 0.98 \\
    Policy entropy coefficient $\eta$ & $3 \times 10^{-4}$ \\
    $\lambda$-return normalization & 95th--5th percentile range \\
    Optimizer eps & $10^{-5}$ \\
    UniMix~\cite{hafner_mastering_2023} regularization & 1\% (for encoder and predictor) \\
    \hline
    \end{tabular}
\end{table}

\bibliography{main}

\end{document}